# The Kinematic Analysis of a Symmetrical Three-Degree-of-Freedom Planar Parallel Manipulator

Damien Chablat and Philippe Wenger
Institut de Recherche en Communications et Cybernétique de Nantes[1]
1, rue de la Noë, 44321 Nantes, France
Damien.Chablat@irccyn.ec-nantes.fr

*Abstract*— Presented in this paper is the kinematic analysis of a symmetrical three-degree-of-freedom planar parallel manipulator. In opposite to serial manipulators, parallel manipulators can admit not only multiple inverse kinematic solutions, but also multiple direct kinematic solutions. This property produces more complicated kinematic models but allows more flexibility in trajectory planning. To take into account this property, the notion of aspects, i.e. the maximal singularity-free domains, was introduced, based on the notion of working modes, which makes it possible to separate the inverse kinematic solutions. The aim of this paper is to show that a non-singular assembly-mode changing trajectory exist for a symmetrical planar parallel manipulator, with equilateral base and platform triangle.

## I. INTRODUCTION

FOR two decades, parallel manipulators have attracted the attention of more and more researchers who consider them as valuable alternative for robotic mechanisms [1-3]. As stated by a number of authors [4], conventional serial kinematic machines have already reached their dynamic performance limits, which are bounded by high stiffness of the machine components required to support sequential joints, links and actuators. Thus, while having good operating characteristics (large workspace, high flexibility and manoeuvrability), serial manipulators have disadvantages of low precision, low stiffness and low power. Also, they are generally operated at low speed to avoid excessive vibration and deflection.

Conversely, parallel kinematic machines offer essential advantages over their serial counterparts (lower moving masses, higher rigidity and payload-to-weight ratio, higher natural frequencies, better accuracy, simpler modular mechanical construction, possibility to locate actuators on the fixed base) that should lead to higher dynamic capabilities. However, most existing parallel manipulators have limited and complicated workspace with singularities, and highly non-isotropic input/output relations [5]. Hence, the performances may significantly vary over the workspace and depend on the direction of the motion.

A well-known feature of parallel manipulators is the existence of multiple direct kinematic solutions (or assembly modes). That is, the mobile platform can admit several positions and orientations (or configurations) in the workspace for one given set of input joint values [6]. The dual problem arises in serial manipulators, where several input joint values correspond to one given configuration of the end-effector. To cope with the existence of multiple inverse kinematic solutions in serial manipulators, the notion of *aspects* was introduced [7]. The aspects were defined as the maximal singularity-free domains in the joint space. For usual industrial serial manipulators, the aspects were found to be the maximal sets in the joint space where there is only one inverse kinematic solution. Many other serial manipulators, referred to as *cuspidal* manipulators, were shown to be able to change solution without passing through a singularity, thus meaning that there is more than one inverse kinematic solution in one aspect. New uniqueness domains have been characterized for cuspidal manipulators [8], [9].

A definition of the notion of aspect was given by [10] for parallel manipulators with only one inverse kinematic solution. These aspects were defined as the maximal singularity-free domains in the workspace. A second definition was given by [11] for parallel manipulators with several inverse kinematic solutions. These aspects were defined as the maximal singularity-free domains in the Cartesian product of the workspace with the joint space.

However, it was shown in [12] that it is possible to link several direct kinematic solutions without meeting a singularity, thus meaning that there exists *cuspidal* parallel manipulators. This property was found for particular links lengths. However, [13] conjectured that such properties cannot exist for symmetrical parallel manipulator. The aim of this paper is to show that a symmetrical 3-DOF planar parallel manipulator can change assembly mode without meeting a singularity. We mean by symmetrical, a manipulator with equilateral base and platform triangles.

This paper is organized as follows. Section II describes the planar 3-RRR parallel manipulator studied, which is used all along this paper. Section III recalls the notion of aspect for parallel manipulators. A non-singular assembly-mode changing trajectory is shown for the symmetrical planar parallel manipulator. The workspace and the generalized

---

aspects are calculated using octree models.

## II. PRELIMINARIES

### A. Parallel manipulator studied

The manipulator under study is a planar three-dof manipulator comprising three parallel RRR chains shown in Fig. 1. This manipulator is used to illustrate the example in this paper. This manipulator has frequently studied, in particular in [6-15].

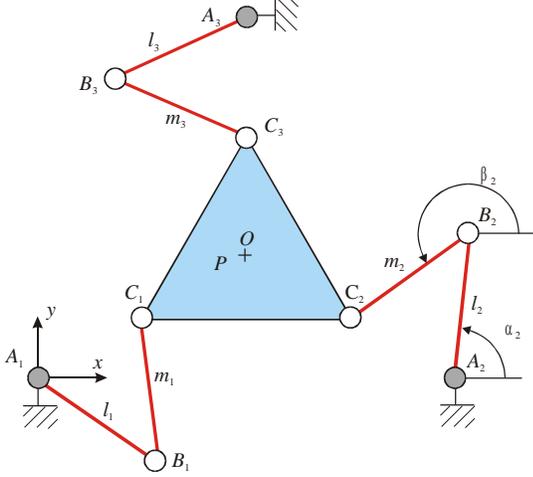

Figure 1: The 3-RRR parallel manipulator studied

The actuated joint variables are the rotation of the three revolute joints located on the base ($\alpha_1, \alpha_2, \alpha_3$). The Cartesian variables are the coordinate $(x, y)$ of the operation point $P$ and the orientation $\theta$ of the platform. The passive and actuated joints will always be assumed unlimited in this study. Points $A_1$, $A_2$ and $A_3$, (respectively $C_1$, $C_2$ and $C_3$) lie at the corners of an equilateral triangle, whose geometric center is $O$ (respectively $P$). Moreover, $l = l_1 = l_2 = l_3 = 6$, with $l_i$ denoting the length of $A_iB_i$, $m = m_1 = m_2 = m_3 = 6$, with $m_i$ denoting the length of $B_iC_i$, $r = r_1 = r_2 = r_3 = 10$, with $r_i$ denoting the length of $A_iO_i$ and $s = s_1 = s_2 = s_3 = 5$, with $s_i$ denoting the length of $C_iP$, in units of length that need not be specified in the paper.

### B. Kinematic Relations

The velocity $\dot{\mathbf{p}}$ of point $P$ can be obtained in three different forms, depending on which leg is traversed, namely,

$$\dot{\mathbf{p}} = \dot{\alpha}_i \mathbf{E}(\mathbf{b}_i - \mathbf{a}_i) + \dot{\beta}_i \mathbf{E}(\mathbf{c}_i - \mathbf{b}_i) + \dot{\theta} \mathbf{E}(\mathbf{p} - \mathbf{c}_i), \quad i \in [1,3] \quad (1)$$

with matrix $\mathbf{E}$,

$$\mathbf{E} = \begin{bmatrix} 0 & -1 \\ 1 & 0 \end{bmatrix}$$

We would like to eliminate the three idle joint rates $\dot{\beta}_1$, $\dot{\beta}_2$ and $\dot{\beta}_3$ from eq. (1), which we do by dot-multiplying eq. (1) by $(\mathbf{c}_i - \mathbf{b}_i)^T$, thus obtaining

$$(\mathbf{c}_i - \mathbf{b}_i)^T \dot{\mathbf{p}} = (\mathbf{c}_i - \mathbf{b}_i)^T \dot{\alpha}_i \mathbf{E}(\mathbf{b}_i - \mathbf{a}_i) + (\mathbf{c}_i - \mathbf{b}_i)^T \dot{\theta} \mathbf{E}(\mathbf{p} - \mathbf{c}_i),$$
$$i \in [1,3] \quad (2)$$

Equation (2) can now be cast in vector form, namely,

$$\mathbf{At} = \mathbf{B}\dot{\boldsymbol{\rho}} \text{ with } \mathbf{t} = \begin{bmatrix} \dot{\mathbf{p}} & \dot{\theta} \end{bmatrix}^T \text{ and } \dot{\boldsymbol{\rho}} = \begin{bmatrix} \dot{\alpha}_1 & \dot{\alpha}_2 & \dot{\alpha}_3 \end{bmatrix}^T \quad (3)$$

with $\dot{\boldsymbol{\rho}}$ thus being the vector of actuated joint rates. Moreover, $\mathbf{A}$ and $\mathbf{B}$ are, respectively, the direct-kinematics and the inverse-kinematics matrices of the manipulator, defined as

$$\mathbf{A} = \begin{bmatrix} (\mathbf{c}_1 - \mathbf{b}_1)^T & -(\mathbf{c}_1 - \mathbf{b}_1)^T \mathbf{E}(\mathbf{p} - \mathbf{c}_1) \\ (\mathbf{c}_2 - \mathbf{b}_2)^T & -(\mathbf{c}_2 - \mathbf{b}_2)^T \mathbf{E}(\mathbf{p} - \mathbf{c}_2) \\ (\mathbf{c}_3 - \mathbf{b}_3)^T & -(\mathbf{c}_3 - \mathbf{b}_3)^T \mathbf{E}(\mathbf{p} - \mathbf{c}_3) \end{bmatrix} \quad (4a)$$

$$\mathbf{B} = \begin{bmatrix} (\mathbf{c}_1 - \mathbf{b}_1)^T \mathbf{E}(\mathbf{b}_1 - \mathbf{a}_1) & 0 & 0 \\ 0 & (\mathbf{c}_2 - \mathbf{b}_2)^T \mathbf{E}(\mathbf{b}_2 - \mathbf{a}_2) & 0 \\ 0 & 0 & (\mathbf{c}_3 - \mathbf{b}_3)^T \mathbf{E}(\mathbf{b}_3 - \mathbf{a}_3) \end{bmatrix} \quad (4b)$$

### C. Singularities

For the planar manipulator studied, such configurations are reached whenever the axes $B_1C_1$, $B_2C_2$ and $B_3C_3$ intersect (possibly at infinity), as depicted in Fig. 2. In the presence of such configurations, the manipulator cannot resist any torque applied at the intersection point $I$.

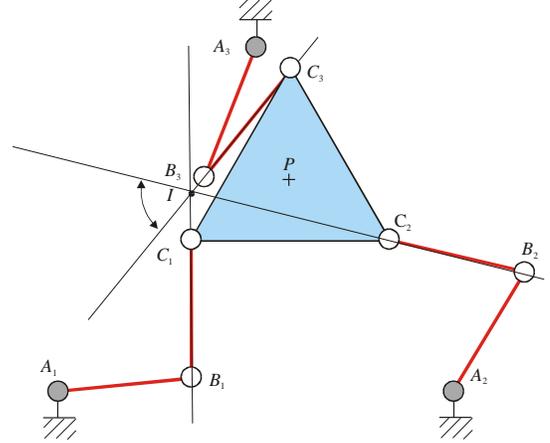

Figure 2: Example of parallel singularity

For the manipulator under study, the serial singularities occur whenever $(\mathbf{b}_i - \mathbf{a}_i)^T (\mathbf{c}_i - \mathbf{b}_i) = l_i m_i$ for at least one value of i, as depicted in Fig. 3 for $i = 1$, i.e. whenever the points $A_i$, $B_i$, and $C_i$ are aligned.

### D. Working modes

The notion of working modes was introduced in [11] for parallel manipulators with several solutions to the inverse kinematic problem and whose matrix $\mathbf{B}$ is diagonal.

A *working mode*, denoted $Mf_i$, is the set of mechanism configurations for which the sign of $B_{jj}$ ($j = 1, \ldots, n$ for a parallel manipulator with $n$ degrees of freedom) does not change and $B_{jj}$ does not vanish. A mechanism configuration is represented by the vector $(X, q)$, which permits us to locate the mobile platform as well as all the links.

$$Mf_i = \left\{ \begin{array}{l} (X, q) \in W \times Q \text{ such that sign}(B_{jj}) = \text{cst for } j = 1, \ldots, n \\ \text{and det}(B) \neq 0 \end{array} \right\}$$

Therefore, the set of working modes ($Mf_i$, $j \in I$) is obtained while using all permutations of sign of each term $B_{jj}$.

The manipulator under study has eight working modes, as depicted in Fig. 4, that we call now (a), (b), ..., (h). Each working mode is defined according to the sign of $B_{jj}$ as is given is in table 1.

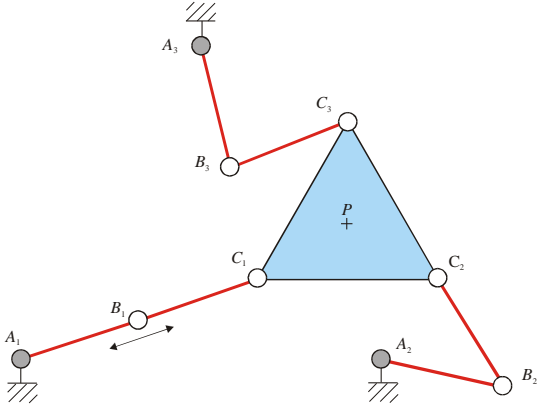

Figure 3: Example of serial singularity when $A_1$, $B_1$ and $C_1$ are aligned

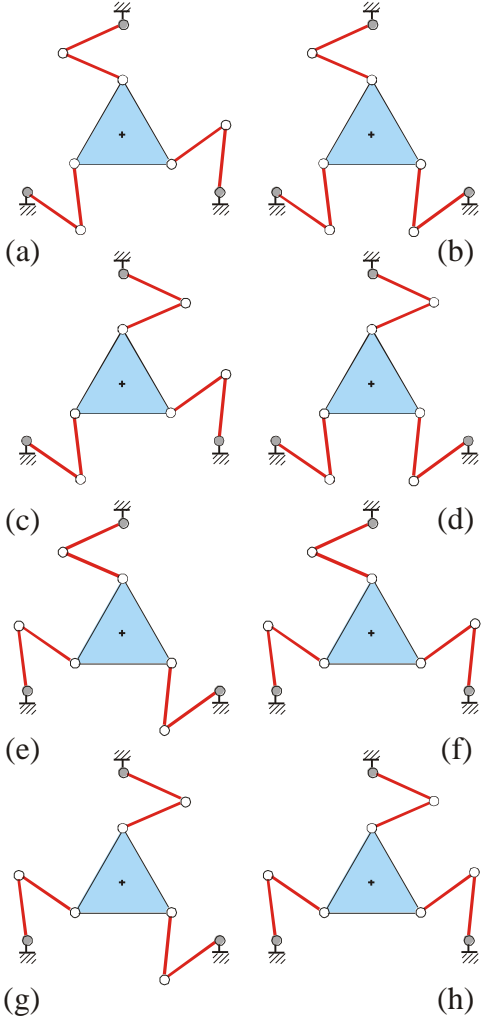

Figure 4: The eight working modes

| Figure 4 | (a) | (b) | (c) | (d) | (e) | (f) | (g) | (h) |
|---|---|---|---|---|---|---|---|---|
| $B_{11}$ | P | P | P | P | N | N | N | N |
| $B_{22}$ | P | N | P | N | N | P | N | P |
| $B_{33}$ | P | P | N | N | P | P | N | N |

Table 1: The eight working modes of the manipulators studied with N (resp. P) denoted negative values of $B_{ii}$ (resp. positive values)

According to each working mode, the parallel singularity locus changes in the workspace, as shown in Fig. 5. In generally, the ability of a parallel manipulator to change its inverse kinematic solution depends on the bounds in the passive and actuated joints. This problem is not taken into account in our study since unlimited joints are assumed.

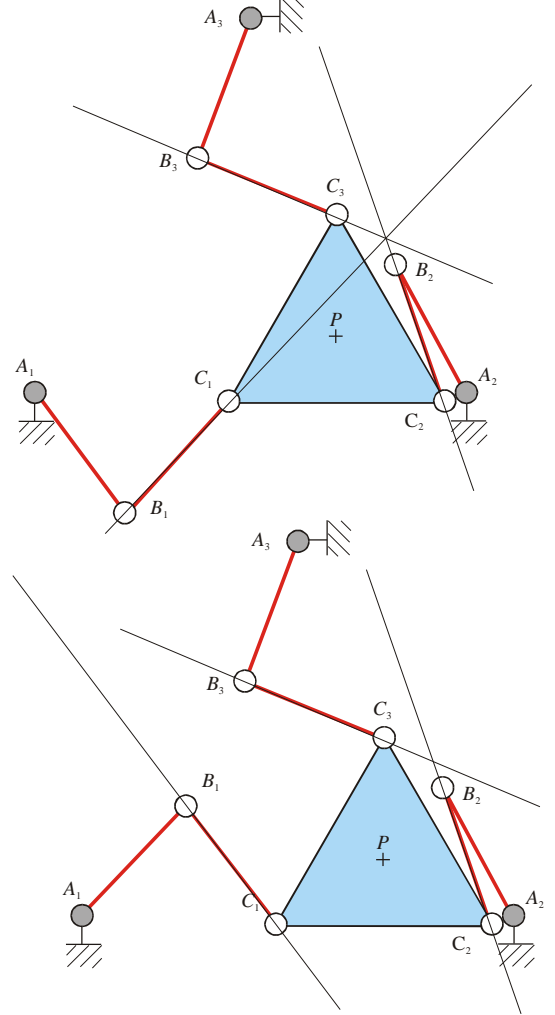

Figure 5: The same platform configuration with two joint configurations (singular on the left and none singular on the right)

*E. Octree Models*

Octree models are hierarchical data structures based on recursive subdivision of the space, respectively [16]. They are useful for representing complex 3 dimensional shapes like workspaces [10]. A close method is used in [17] that divides the workspace into boxes. This method does not use recursive subdivision but interval analysis methods [18] to build the dextrous workspace. However, it does not make it possible to perform Boolean operations or to make path-connectivity analysis easily. The first method permits us to calculate easily all kind of space and the computing time is limited as a function of the accuracy. The second one is more exact but requires more ability to be implemented. In both cases, we can characterize spaces whose dimensions are either lengths or angles.

Since the structure of the octree model has an implicit adjacency graph, path-connectivity analyses and trajectory planning can be carried out naturally. The optimal construction method of a $2^k$-tree is derived from the shape, which recalls the tree. The most interesting approach consists in testing successively all the nodes present in the maximal depth, following an order of numbering which quickly allows nodes to be grouped and thus, simplifies the $2^k$-tree. The order of numbering for this algorithm is based on Morton's sweeping [19]. The inverse or direct kinematic model is used to calculate $2^k$-tree.

The figure 6 represents the octree model of its Cartesian workspace where the first and the second axis represent the position and the third axis the orientation of the mobile platform.

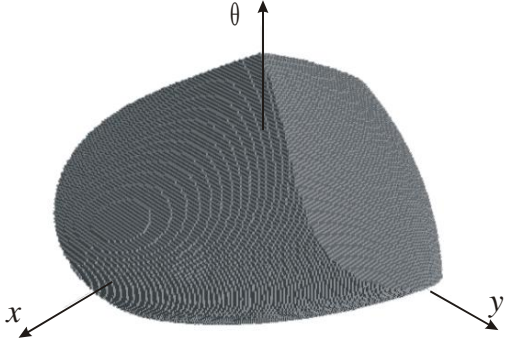

Figure 6: The Cartesian workspace

### III. WORKSPACE ANALYSIS

#### A. Aspect definitions

The notion of aspect was introduced by [7] to cope with the existence of multiple inverse kinematic solutions in serial manipulators. Recently, the notion of aspect was defined for parallel manipulators with only one inverse kinematic solution to cope with the existence of multiple direct kinematic solutions [10] and for parallel manipulators with multiple inverse and direct kinematic solutions (the *generalized aspects* [11]). For the manipulator studied, we use the second definition.

The generalized aspects $A_{ij}$ are defined as the maximal sets in $W \times Q$ so that
- $A_{ij} \subset W \times Q$;
- $A_{ij}$ is connected;
- $A_{ij} = \{(X, q) \in Mf_i$ such that $\det(\mathbf{A}) \neq 0\}$

In other words, the generalized aspects $A_{ij}$ are the maximal singularity-free domains of the Cartesian product of the reachable workspace (called *W*) with the reachable joint space (called *Q*).

The projection of the generalized aspects onto the workspace yields the parallel aspects $WA_{ij}$ so that,
- $WA_{ij} \subset W$;
- $WA_{ij}$ is connected.

The parallel aspects are the maximal singularity-free domains in the workspace for one given working mode.

The projection of the generalized aspects onto the joint space yields the serial aspects $QA_{ij}$ so that,

- $QA_{ij} \subset Q$;
- $QA_{ij}$ is connected.

The serial aspects are the maximal singularity-free domains in the joint space for one given working mode.

For each working mode, there exists, at least, one aspect where $\det(\mathbf{A})$ is positive and another one where $\det(\mathbf{A})$ is negative. However, such regions can be disjoint. In table 2, we associated the aspects with a working mode for which $\det(\mathbf{A})$ is positive. For the working mode (a), there exist four aspects and for the other ones, there is only one aspect. Due to the symmetrical properties of the mechanism, there exist also 11 aspects where $\det(\mathbf{A})$ is negative.

| Working modes | (a) | (b) | (c) | (d) | (e) | (f) | (g) | (h) |
|---|---|---|---|---|---|---|---|---|
| N° figure | 7 | 8 | 9 | 10 | 11 | 12 | 13 | 14 |

Table 2: The projection of the generalized aspects on the workspace when $\det(\mathbf{A}) > 0$ for each working mode

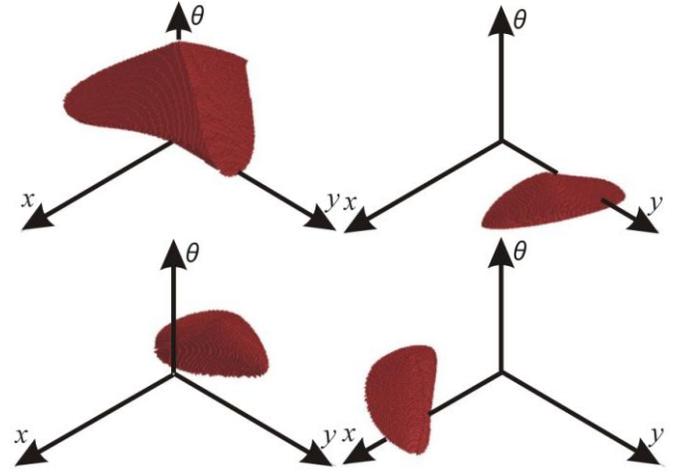

Figure 7: The four parallel aspects for the working mode (a) and $\det(\mathbf{A}) > 0$

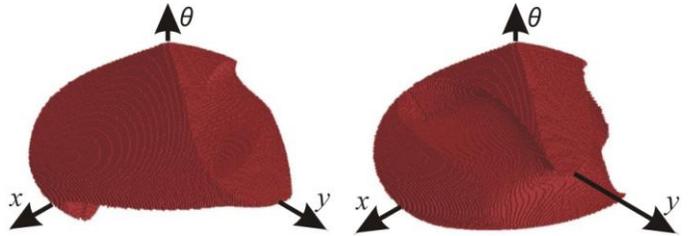

Figure 8: The parallel aspect for the working mode (b) and $\det(\mathbf{A}) > 0$    Figure 9: The parallel aspect for the working mode (c) and $\det(\mathbf{A}) > 0$

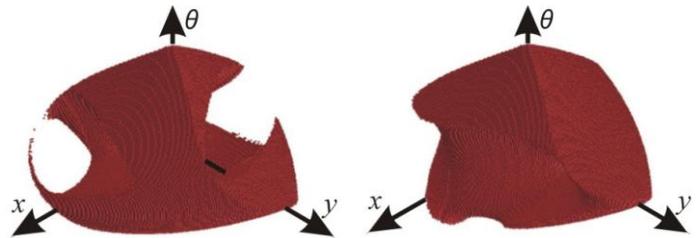

Figure 10: The parallel aspect for the working mode (d) and $\det(\mathbf{A}) > 0$    Figure 11: The parallel aspect for the working mode (e) and $\det(\mathbf{A}) > 0$

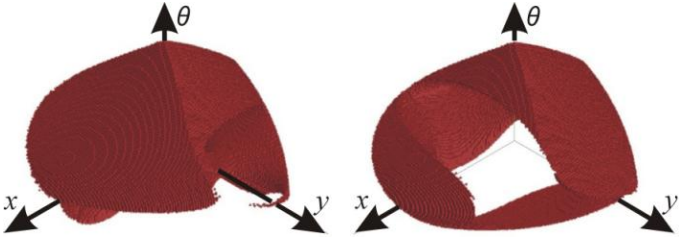

Figure 12: The parallel aspect for the working mode (f) and $\det(\mathbf{A}) > 0$

Figure 13: The parallel aspect for the working mode (g) and $\det(\mathbf{A}) > 0$

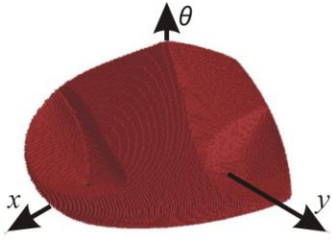

Figure 14: The parallel aspect for the working mode (h) and $\det(\mathbf{A}) > 0$

The calculation of the generalized aspects can be performed by using a $2^6$ octree model or two octree models. We use the second method. The first one is the projection of the generalized aspect onto the reachable workspace and the second one the projection onto the reachable joint space for a given working mode and a given sign of $\det(\mathbf{A})$ constant. To obtain these results with an accuracy of 0.093 for the position and 1,4 degrees for the orientation, the computing times is 90 seconds with an AMD Athlon XP processor 2500$^+$ and the maximum memory used is 180 Mb. The connectivity analysis of each domain requires 20 seconds.

### B. Non-singular posture changing trajectories

In [12], a non-singular posture changing trajectory was found for a 3-RPR planar manipulator. However, it appears that this trajectory passes close to a singular configuration. This property was confirmed in [10] for the same manipulator and for a 3-RRR planar manipulator with non-symmetrical geometry [20].

According to the assumption in [13], we can think that such properties may not exist for the mechanism studied. However, we shown that a non-singular configuration changing trajectories exists.

For the following input joint values:

$\theta_1 = 5.862610$, $\theta_2 = 1.277470$, $\theta_3 = 5.213885$

Four direct kinematic solutions are found (Figure 15 and Table 3). We notice that solutions 1 and 4 are in the same generalized aspect (The parallel aspect associated is depicted in the figure 8).

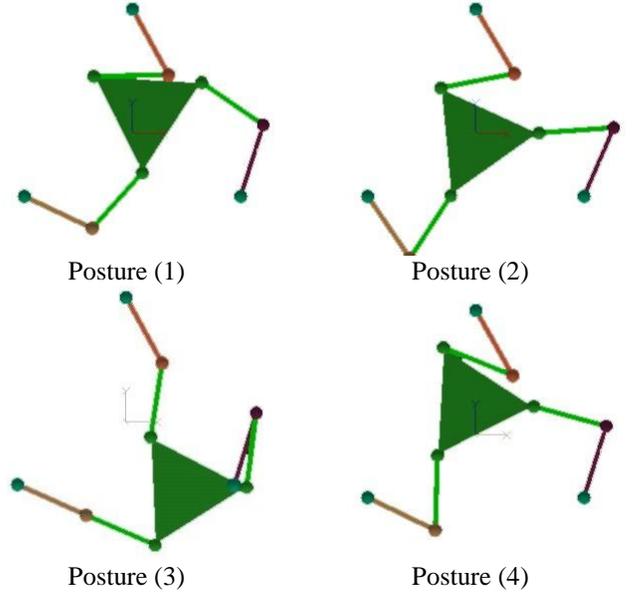

Posture (1)     Posture (2)

Posture (3)     Posture (4)

Figure 15: The four direct kinematic solutions for $\theta_1 = 5.862610$, $\theta_2 = 1.277470$, $\theta_3 = 5.213885$

| Posture N° | x | y | θ in degrees |
|---|---|---|---|
| (1) | 1.102 | 1.956 | 57.50 |
| (2) | 0.705 | 2.751 | 46.85 |
| (3) | 4.638 | -5.413 | 32.35 |
| (4) | -0.357 | 2.720 | 26.51 |

Table 3: Four direct kinematic solutions for the same joint values

A first method to confirm this property is to evaluate the determinant of $\mathbf{A}$ and $B_{ii}$ (Table 4) and to find out a trajectory between these two postures.

|  | det($\mathbf{A}$) | $B_{11}$ | $B_{22}$ | $B_{33}$ |
|---|---|---|---|---|
| Posture (1) | 307.990 | -34.132 | -34.008 | 31.827 |
| Posture (4) | 522.868 | -33.023 | -35.997 | 21.7203 |

Table 4: Evaluation of $\det(\mathbf{A})$ and $B_{ii}$ for the two pose in the same generalized aspect

We find a non-singular continuous trajectory between postures (1) and (4) by passing through an intermediate posture (5) (Figure 16) whose position is (-0.987; 1.930) and orientation is 12.35 degrees. Between these three postures, a linear interpolation is defined to stay in the same generalized aspect. The values of $\det(\mathbf{A})$, $B_{11}$, $B_{22}$ and $B_{33}$ are evaluated and each value of these indices is normalized by its maximum value as it is shown in figure 17.

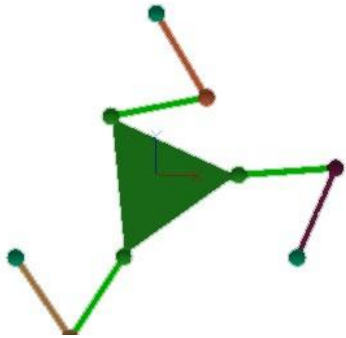

Figure 16: The intermediate posture (5) for the non-singular changing trajectory

With this result, we have proofed that a non-singular assembly mode trajectory is possible for a symmetrical planar 3-RRR parallel manipulator. Inside such trajectory, not any kinematic index, derived from the Jacobian matrices, permits us to recognize such property.

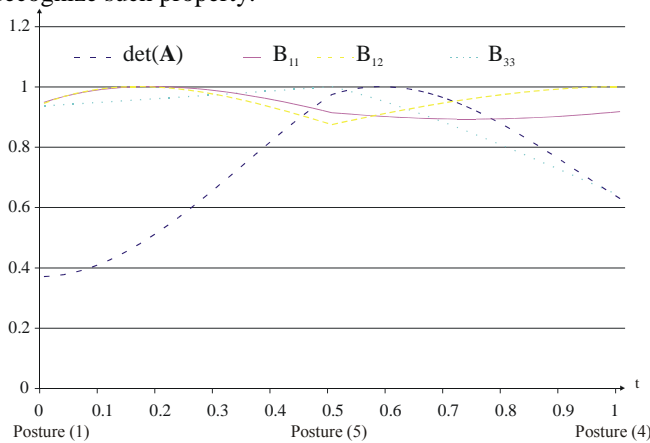

Figure 17: Variations of the normalized values of det(A), $B_{11}$, $B_{22}$ and $B_{33}$ along the trajectory (t) between postures (1) and (4)

### C. Characteristic surfaces

To separate the direct and inverse kinematic solutions, the uniqueness domains are determined for the parallel manipulator with one inverse kinematic solution in [10] and for parallel manipulator with several inverse kinematic solutions in [20]. The boundaries of the uniqueness domains are defined by the characteristic surfaces [20]. For the generalized aspect (b), we can compute the characteristic surface that permits us to isolate the assembly modes where it is possible to realize non-singular assembly mode changing trajectories (Figure 18).

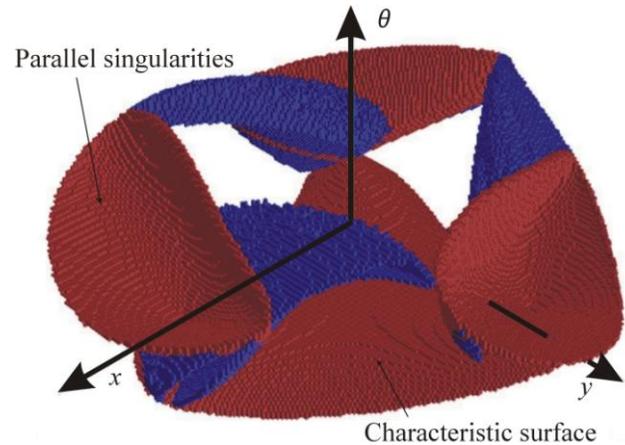

Figure 18: The parallel singularities and the characteristic surfaces associated with the generalized aspect (b)

### IV. SUMMARY AND CONCLUSIONS

A kinematic analysis of a planar 3-RRR parallel manipulator with symmetrical properties was presented in this paper. The eight working modes have been characterized and 22 generalized aspects have been found out. Inside such domains, any continuous trajectories are possible. In such domains, there are non-singular changing trajectories but not any kinematic index can recognize such property. An example of non-singular changing trajectory is given and the characteristic surface are computed which permit, in a future works, to define closely the uniqueness domains of the manipulator studied.

### REFERENCES

[1] H. Asada and J.J. Slotine, "Robot Analysis and Control," John Wiley & Sons, (1986).
[2] K.S. Fu, R. Gonzalez and C.S.G. Lee, "Robotics: Control, Sensing, Vision, and Intelligence," McGraw-Hill, (1987).
[3] J.J. Craig, "Introduction to Robotics: Mechanics and Control," Addison Wesley, (1989).
[4] L.W. Tsai, "Robot Analysis, The Mechanics of Serial and Parallel Manipulators," John Wiley & Sons, (1999).
[5] D. Stewart, "A Platform with Six Degrees of freedom," Proceedings of the Institution of Mechanical Enginners Vol. 180, Part 1, No. 15, 371-386, (1965).
[6] J-P. Merlet, "Parallel robots," Kluwer Academic Publ., Dordrecht, The Netherland, (2000).
[7] P. Borrel, "A study of manipulator inverse kinematic solutions with application to trajectory planning and workspace determination," Proc. IEEE Int. Conf on Rob. And Aut., pp 1180-1185, (1986).
[8] Ph. Wenger, "A new general formalism for the kinematic analysis of all nonredundant manipulators," IEEE Robotics and Automation, pp. 442-447, (1992).
[9] J. El Omri, "Analyse Géométrique et Cinématique des Mécanismes de Type Manipulateur," Thèse, Nantes, (1996).
[10] Ph. Wenger and D. Chablat, "Uniqueness Domains in the Workspace of Parallel Manipulators," IFAC-SYROCO, Vol. 2, pp. 431-436, 3-5 Sept., Nantes, (1997).

[11] D. Chablat and Ph. Wenger, "Working Modes and Aspects in Fully-Parallel Manipulator," IEEE International Conference on Robotics and Automation, pp. 1970-1976, May, (1998).

[12] C. Innocenti and V. Parenti-Castelli, "Singularity-free evolution from one configuration to another in serial and fully-parallel manipulators," Robotics, Spatial Mechanisms and Mechanical Systems, ASME, (1992).

[13] P.R. McAree, R. W. Daniel, "An explanation of Never-Special Assembly Changing Motion for 3-3 Parallel Manipulators," Int. Journal of Robotics research, Vol. 18/6, June (1999).

[14] J. Angeles, "Fundamentals of Robotic Mechanical Systems," Springer-Verlag, (2002).

[15] C. Gosselin, and J. Angeles, "The Optimum Kinematic Design of a Planar Three-Degree-of-Freedom Parallel Manipulator," ASME, Journal of Mechanisms, Transmissions, and Automation in Design, Vol. 110, March (1988).

[16] D. Meagher, "Geometric Modelling using Octree Encoding," Technical Report IPL-TR-81-005, Image Processing Laboratory, Rensselaer Polytechnic Institute, Troy, New York 12181 (1981).

[17] D. Chablat, Ph. Wenger and J-P. Merlet, "Workspace Analysis of the Orthoglide using Interval Analysis," 8th International Symposium on Advances in Robot Kinematics, Kluwer Academic Publishers, Caldes de Malavella, Espagne, June (2002).

[18] J-P. Merlet, "ALIAS: an interval analysis based library for solving and analyzing system of equations," Séminaire Systèmes et équations algébriques, Toulouse, pp. 1964-1969, June (2002).

[19] G. Morton, "A Computer Oriented Geodetic Data Base and a new Technique in File Sequencing," IBM Ltd, Ottawa, Canada (1966).

[20] D. Chablat and Ph. Wenger, "Les Domaines d'Unicité des Manipulateurs Pleinement Parallèles," Journal of Mechanism and Machine Theory, Vol 36/6, pp. 763-783, (2001).